\documentclass[twoside,leqno,twocolumn]{article}

\usepackage[letterpaper]{geometry}

\usepackage{ltexpprt}

\usepackage{cite}
\usepackage{textcomp}
\usepackage{xcolor}
\usepackage{dsfont}

\usepackage{amsmath,amsbsy,amsfonts,amssymb,amsthm,units,bm}
\usepackage[mathscr]{eucal}
\usepackage[algo2e]{algorithm2e} 
\usepackage{algorithm}

\usepackage{graphicx}
\usepackage{tikz}
\usepackage{stackengine}
\usepackage{subfigure}
\usepackage{subfiles}

\theoremstyle{definition}

\makeatletter
\newsavebox\myboxA
\newsavebox\myboxB
\newlength\mylenA

\newcommand*\xbar[2][0.75]{%
    \sbox{\myboxA}{$\m@th#2$}%
    \setbox\myboxB\null% Phantom box
    \ht\myboxB=\ht\myboxA%
    \dp\myboxB=\dp\myboxA%
    \wd\myboxB=#1\wd\myboxA% Scale phantom
    \sbox\myboxB{$\m@th\overline{\copy\myboxB}$}%  Overlined phantom
    \setlength\mylenA{\the\wd\myboxA}%   calc width diff
    \addtolength\mylenA{-\the\wd\myboxB}%
    \ifdim\wd\myboxB<\wd\myboxA%
       \rlap{\hskip 0.5\mylenA\usebox\myboxB}{\usebox\myboxA}%
    \else
        \hskip -0.5\mylenA\rlap{\usebox\myboxA}{\hskip 0.5\mylenA\usebox\myboxB}%
    \fi}
\makeatother

\newcommand{\tens}[1]{\boldsymbol{\mathscr{#1}}}
\newcommand{\vect}[1]{\ensuremath{\mathbf{#1}}}

\newcommand{\argmax}{\mathop{\rm argmax}}

\newcommand{\tB}{\tens{B}}

\newcommand{\tS}{\tens{S}}

\newcommand{\tD}{\tens{D}}

\newcommand{\w}{\bm{\omega}}

\newcommand{\x}{\vect{x}}

\newcommand{\y}{\vect{y}}

\providecommand{\keywords}[1]{\textbf{\textit{Keywords---}} #1}

\begin{document}

%\setcounter{chapter}{2} % If you are doing your chapter as chapter one,
%\setcounter{section}{3} % comment these two lines out.

% \title{\Large SIAM/ACM Preprint Series Macros for
% Use With LaTeX\thanks{Supported by GSF grants ABC123, DEF456, and GHI789.}}
% \author{Corey Gray\thanks{Society for Industrial and Applied Mathematics.}
% \and Tricia Manning\thanks{Society for Industrial and Applied Mathematics.}}

\title{SemiFed: Semi-supervised Federated Learning with Consistency and Pseudo-Labeling}

\author{\fontsize{10.8pt}{\baselineskip}\selectfont Haowen Lin\textsuperscript{1}, Jian Lou\textsuperscript{2,3,}\thanks{Corresponding Author.}~, Li Xiong\textsuperscript{2}, Cyrus Shahabi\textsuperscript{1}\\
\fontsize{10pt}{\baselineskip}\selectfont \textsuperscript{1}University of Southern California\ \ \ \textsuperscript{2}Emory University\ \ \ \textsuperscript{3}Xidian University\\
{\tt\small haowenli@usc.edu\ \ jlou@xidian.edu.cn\ \ lxiong@emory.edu\ \ shahabi@usc.edu}
}

\maketitle

% Copyright Statement
% When submitting your final paper to a SIAM proceedings, it is requested that you include 
% the appropriate copyright in the footer of the paper.  The copyright added should be 
% consistent with the copyright selected on the copyright form submitted with the paper.
% Please note that "20XX" should be changed to the year of the meeting.

% Default Copyright Statement
\fancyfoot[R]{\scriptsize{Copyright \textcopyright\ 2021 by SIAM\\
Unauthorized reproduction of this article is prohibited}}

% Depending on which copyright you agree to when you sign the copyright form, the copyright 
% can be changed to one of the following after commenting out the default copyright statement
% above.

%\fancyfoot[R]{\scriptsize{Copyright \textcopyright\ 20XX\\
%Copyright for this paper is retained by authors}}

%\fancyfoot[R]{\scriptsize{Copyright \textcopyright\ 20XX\\
%Copyright retained by principal author's organization}}

%\pagenumbering{arabic}
%\setcounter{page}{1}%Leave this line commented out.

\begin{abstract} \small\baselineskip=9pt 
Federated learning enables multiple clients, such as mobile phones and organizations, to collaboratively learn a shared model for prediction while protecting local data privacy. However, most recent research and applications of federated learning assume that all clients have fully labeled data, which is impractical in real-world settings. In this work, we focus on a new scenario for cross-silo federated learning, where data samples of each client are partially labeled. We borrow ideas from semi-supervised learning methods where a large amount of unlabeled data is utilized to improve the model's accuracy despite limited access to labeled examples. We propose a new framework dubbed SemiFed that unifies two dominant approaches for semi-supervised learning: consistency regularization and pseudo-labeling. SemiFed first applies advanced data augmentation techniques to enforce consistency regularization and then generates pseudo-labels using the model's predictions during training. SemiFed takes advantage of the federation so that for a given image, the pseudo-label holds only if multiple models from different clients produce a high-confidence prediction and agree on the same label. Extensive experiments on two image benchmarks demonstrate the effectiveness of our approach under both homogeneous and heterogeneous data distribution settings.\end{abstract}


\section{Introduction}

While deep neural networks have shown promising results in various computer vision applications and tasks, their success is largely
attributed to the availability of a significant amount of training data \cite{mahajan2018exploring,caron2018deep,locatello2019challenging}.   Fortunately, with the rapid proliferation of smart devices and sensors, multiple clients such as mobile device users or organizations can collaboratively learn a global model via communication, thus alleviating the burden on data gathering for a single entity. Traditional machine learning algorithms require centralizing all data samples in a single machine or data center, which is impractical due to privacy concerns, high storage costs, and legal constraints. This motivates federated learning (FL) \cite{mcmahan2017communication}, which is a promising machine learning paradigm in which a loose federation of clients participate in collaborative training under the coordination of a central server.

Although federated learning has attracted much attention in both academia and industry, existing FL applications and approaches mainly focus on fully supervised settings where all the input data have labels. However, this assumption is not practical, because manually adding high-quality labels to all training data may not be equally feasible by all clients. A relevant approach in centralized learning is semi-supervised learning (SSL), which leverages unlabeled samples in addition to a small portion of labeled examples to obtain performance gains. However, it is not trivial to marry centralized SSL techniques with FL training due to its distinctive requirements, rendering a direct application of SSL impractical for real FL settings.

% Moreover, since the server is not trustworthy and can mishandle this sensitive information, many recent studies \cite{abadi2016deep,shokri2015privacy} have 
% proposed to advanced privacy-preserving mechanisms such as Differential Privacy (DP)\cite{Dwork2011} to further prevent leaking sensitive information to the central server.

One of the most challenging problems in FL setting is that data among clients can be non-independent and identically (non-IID). In this case, even labeled and unlabeled data may come from different underlying distributions, and the unlabeled data may contain classes not present in the labeled data. This non-IID situation can have a paramount impact on the SSL algorithm design. For example, \cite{oliver2018realistic} shows that even in centralized setting, utilizing data from a mismatched set of classes can even hurt the performance of many SSL methods compared to not using any unlabeled data at all. While \cite{oliver2018realistic} synthetically varies the class overlap, this mismatch between labeled and unlabeled categories naturally arises under the non-IID data partitions in FL setting. Therefore, it is important to design a method that is not sensitive to the non-IID data distributions and does not deteriorate by adding unlabeled data from a mismatched set of classes.

\begin{figure}[]
\centering
  \includegraphics[width=.98\linewidth]{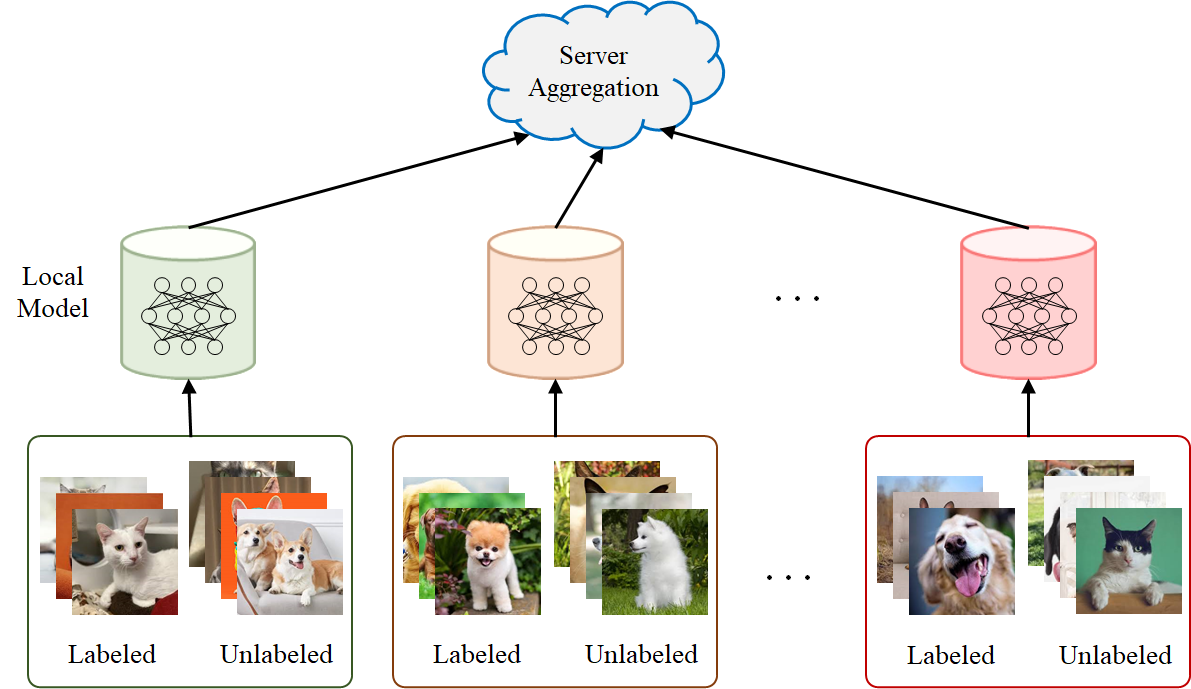}
  \caption{Illustration of a realistic Federated Semi-supervised learning scenario}
    \label{img:scenario}
\end{figure}

We propose Semi-Supervised Federated Learning (SemiFed) as a unified framework and apply it to image classification with limited labeled samples. Figure \ref{img:scenario} presents the illustration of the limited labeled data under collaborative training scenario. SemiFed consists of two key components. First, it performs consistency regularization, which encourages the network to produce the same output distribution when its inputs are perturbed and can be applied to all samples without labels. We carefully choose the noise injected into consistency training. We find that advanced data augmentation \cite{cubuk2018autoaugment,cubuk2020randaugment} techniques that boost network accuracy in fully supervised tasks can be leveraged for the perturbations in our FL semi-supervised learning. Second, after several rounds of warm-up training phases, we adopt the idea of pseudo-labeling \cite{lee2013pseudo} which first produces an artificial label for each unlabeled image and then enforces the model to predict the artificial label when fed the unlabeled sample as input in the following training stages. However, training the network with falsely inferred pseudo-labels may degrade the model performance. To solve this issue, we keep the artificial labels only when the local models and the global model assign a very high probability (high confidence) to one of the possible classes. Further, to take advantage of the federation, we transmit local models and assign a pseudo-label to an unlabeled sample only when multiple models agree on the same label candidate with high confidence. With the integration of consistency regularization and federated pseudo-labeling, we conduct experiments to evaluate the proposed approach on standard image benchmarks under federated learning setting. We show that our algorithm is consistent under non-IID distribution data, which is significant under FL setting.

% \begin{enumerate}
% \item We introduce an under-explored yet practical federated learning scenario where the clients' data are only partially labeled, which poses several new challenges to achieve high performance while ensuring privacy restrictions. 
% % We introduce a new challenge in federated learning, which previous work has often ignored, and formally define the scenario that data stored in clients is partially labeled without violating privacy constraints.
% \item We propose SemiFed, a unified framework that tackles limited labeled sampling problem in federated learning based on a thoughtful integration and further algorithmic adaptations specific to the FL setting by including consistency regularization with carefully chosen noise and pseudo-labeling with consensus across models trained on different clients.
% \item We deploy our algorithm in the real distributed training system and conduct experiments to evaluate the proposed approach on standard image benchmarks under federated learning setting to show it outperforms state-of-the-art approaches up to 4.39\% improvement. We show that our algorithm is consistent under non-IID distribution data, which is significant in federated learning.
% \end{enumerate}

The remainder of the paper is organized as follows. Section \ref{sec:background_related_work_section} presents preliminaries of FL and related work. Section \ref{SemiFed_Section} describes
the scenario and methodology of SemiFed. Section \ref{experimenet_section} presents the experiments and evaluation of the results. Finally, Section \ref{conclusion_section} concludes the paper with a discussion of future work.

\section{Background and Related Work}
\label{sec:background_related_work_section}
% We seek to solve the challenge that under federated learning settings when all data are separately stored among different clients, only a few of them are labeled. We first provide background on conventional federated learning and its popular algorithm. Then we will formally define a realistic limited label scenario under federated learning framework.

In this section, we first formally define the problem of SSL in federated learning where clients may only create or obtain a small portion of labels for their local datasets. Then, we present related work that attempts to solve the problem of limited labeled samples in federated learning. 

% \subsection{Federated Learning} Federated learning aims to collaboratively train a machine learning model based on all the training data samples that are stored separately in different clients, while avoiding sharing raw data over the communication network. Here, we review the background on the current federated learning and review its most fundamental algorithm, i.e., FedAvg.
% % However, few works address the challenge of how to leverage unlabeled data when federated learning clients have only a limited number of labeled samples. 

 \emph{1) Problem setting.} Let $\tD$ be the dataset with $C$ categories. We assume that there are $K$ clients (e.g., edge devices) participating in the federated learning. Each client has a dataset $\tD^k$ with $C$ categories that are separated into labeled dataset $\tD_l^k :=\{\x_i,\y_i\}^{N_l^k}_{i=1}$ with $\x_{i}$ denoting the $i$-th sample and $\y_i \in \{1, 2, . . . , C\}$ being the corresponding label, and unlabeled dataset $\tD_u^k := \{\x_j\}^{N_u^k}_{j=1}$. The total number of data samples across all clients is therefore  $N_l = \sum_{i=1}^{K} (N_l^k +N_u^k)$ . $N_l^k$ is the number of labeled data samples for $k$th client, $N_u^k$ is the number of unlabeled data samples, and
usually, the labeled data are much less than unlabeled data, i.e.,
 $N_l^k \ll N_u^k$.  Depending on the data distribution, the local datasets $\tD^k$ could either be homogeneous, i.e. independent and identically distributed (IID)), or heterogeneous with varying distribution from clients to clients (i.e., non-IID). 

\emph{2) Related work.}
Although limited, there still exist some attempts to the limited labeled samples problem in federated learning. Guha et al. \cite{guha2019one} propose one-shot federated learning, where they let the server learn a successful global model over a network of federated devices in a single round of communication in both supervised and semi-supervised learning. In their setting, all clients have labeled data, and only the central server has unlabeled data. Sharma et al. propose to solve federated transfer learning under Secret Sharing protocols. They assume that each client can have different datasets with totally different labels. However, we focus on the scenario when the server is not allowed to have any data for privacy reasons, which is different from the setting introduced in previous work. Lastly, a parallel work that targets this challenge is \cite{jeong2020federated}, where it imposes consistency loss among multiple clients and decomposes model parameters for labeled and unlabeled data to exploit unlabeled information. However, they do not provide detailed experiment results to show model performance under the different number of labeled samples under IID and non-IID settings.

% The whole process is summarized in Algorithm \ref{alg.FedAvg}.

% \begin{algorithm}
% \SetAlgoLined

% Initialize $\w_0$; \;

% \For{$t=0 ,...,T-1$}{
%     Server samples a random sets $\tS$ from $K$ clients\;
    
%     $\w^k_{t,0} = \w_t$;\;
    
%     \For{each client $k \in \tS$ in parallel}{
%         %$\tB \leftarrow $ split local data into batches of size $B$ \; \\
%         \For{local epoch e from $0$ to $E-1$}{
%             \For{iteration i from 0 to $I-1$}
%             {
%              Sample mini-batch $\tB$ from $k$'s local dataset;\;\\
%              $\w_{t,i+1}^k = \w_{t,i}^k - \eta \cdot \newline \frac{1}{|\tB|}\sum _{b\in\tB}\nabla l(\w^k_{t,i};\x_{b},\y_b)$;
%             %$\w_{t}^k \leftarrow \w_{t}^k - \eta \cdot \nabla l(b;w^k_{t})$
%             }
%             %Sample mini-batch $\tB$;\;

%         }

%     }
%     $\w_{t+1} = \frac{1}{|\tS|} \sum\limits_{k \in \tS} \w^k_{t,EI}$;
    
% } 

%  \caption{\textbf{FedAvg}, $|\tB|$ is the local minibatch size, $T$ is the number of communication rounds, and $\eta$ is the learning rate,E is the number
% of local epochs, $I$ is the epoch iterations}
%  \label{alg.FedAvg}
% \end{algorithm}

\section{SemiFed Framework}
\label{SemiFed_Section}

SemiFed unifies two powerful approaches that can leverage the unlabeled data at each local site. First, SemiFed utilizes consistency regularization to leverage unlabeled data. Second, it uses pseudo-labeling to progressively label unlabeled data using the models trained so far to guide the remaining learning process. As a key innovation, we also adapt the pseudo-labeling method to FL setting, where we require multiple local models and the global model to agree on the labels before we make use of the labels for future training.

\subsection{Consistency Regularization}
\label{Consistency Regularization}

Consistency regularization is one of the most significant approaches to learn from unlabeled data and improve the performance of labeled data \cite{bachman2014learning,miyato2018virtual}. In a nutshell, Consistency Regularization utilizes the unlabeled data by enforcing model predictions to be invariant under any kind of small random transformation of the data and perturbations to the model. The design relies on the low-density separation assumption that the decision boundary should not traverse samples of high-density regions but should stay in low-density regions \cite{zhu2005semi}. Thus, the model should be robust to any small changes fed into the input or any hidden layer.

Following the work in \cite{bachman2014learning,miyato2018virtual},  we formulate the objective function to incorporate information from both labeled and unlabeled data at each client. It includes two loss terms: a supervised loss $l_s^k$ applied to labeled data and an unsupervised loss $l_u^k$ applied to data without class label at $k$th client. Specifically, we define our $l_s$ as the standard cross-entropy loss on the labeled examples where $p_w^k$ is the local model in $k$-th client:
\begin{equation}
\label{supervised_loss_equation}
l_s^k = \sum_{i=1}^{N_l^k}-\log p_{\w}^k(\y_i|\x_i).
\end{equation}
For unlabeled data, we focus on the setting where the noise is fed into the input $\x$ (i.e. $\Tilde{\x} = a(\x,\epsilon)$). We represent the distribution of
the current predictions of samples $p_{\w} (\y|\x)$ and generate an auxiliary distribution $p_{w}(\y|\Tilde{\x})$ from the perturbed data $\Tilde{\x}$. Our goal is to minimize the distance between these two distributions using the KL divergence, a widely used
distribution-wise asymmetric distance measure. Therefore, the unsupervised loss for $k$th client is defined as:
\begin{equation}
\label{unsupervised_loss_equation}
l_u^k = \text{KL}(p_{\w}^k (\y|\x) || p_{\w}^k(\y|\Tilde{\x})). 
\end{equation}
For jointly training with all data samples, the full objective function is thus given by
\begin{equation}
l_s^k + \lambda_u l_u^k,
\end{equation}
where $\lambda_u$ is a hyper-parameter that 
controls the balance between the supervised cross entropy and the unsupervised consistency training loss.

Previous works have shown that one effective way to improve the performance of this consistency training procedure is to ``wisely'' choose the type of noise injection \cite{tarvainen2017mean}. A straightforward way is to employ random noise functions such as Gaussian noise and dropout noise in the vision domain \cite{laine2016temporal}. However, simply injecting a random noise cannot effectively push the boundary line to cross the low-density region or can make the predictor unstable under a small perturbation in a specific direction \cite{goodfellow2014explaining}. To alleviate this problem, \cite{miyato2018virtual} searches adversarial perturbations that maximize the change in model prediction. \cite{berthelot2019mixmatch,berthelot2019remixmatch} generates noise by interpolating random unlabeled data points. More recent work shows that using stronger data augmentations such as augmentation policies searched by reinforcement learning not only improves the generalization of deep learning models but also can be beneficial to the objective in consistency learning framework \cite{cubuk2018autoaugment}

Following the idea in \cite{xie2019unsupervised}, we take advantage of the advanced data augmentation technique called RandAugment \cite{cubuk2020randaugment} that is initially inspired by AutoAugment  \cite{cubuk2018autoaugment}. AutoAugment uses a search method to automatically find effective data augmentation strategies that include all transform operations in the Python Image Library (PIL) for a specific dataset. RandAugment does not require a separate search phase on a proxy task but instead uniformly takes operation from the PIL library with similar performance compared with AutoAugment. Because we do not have enough labels for the dataset in our settings, we explore RandAugment that does not require any labeled data to search for the policy. We choose such advanced data augmentation as our noise injection method because it can generate meaningful augmented samples under large modifications to the data input. We show that using RandAugment on the training set yields positive results in the experiment.

\subsection{Pseudo-Labeling}
\label{Pseudo-Labeling}

Our algorithm also incorporates the idea of pseudo-labeling \cite{lee2013pseudo}, a simple but efficient method that can further improve the performance of classifier when we train using the labeled and unlabeled data simultaneously. Pseudo-labeling first trains the model only on  labeled dataset. At each iteration, the model uses predictions from previous iteration as target classes for unlabeled data samples as if they were true labels. The process goes through fixed number of iterations until all unlabeled data are labeled. Pseudo-label advances the self-training in the way that it would use both labeled data and unlabeled
data for training \cite{lee2013pseudo}. For each unlabeled data sample, it would pick the class with maximum predicted confidence as prediction to be used as a pseudo label,
\begin{equation}\label{eqs12}
    {\Tilde{\y}}_c= 
\begin{cases}
    1,  & \text{if } c= \argmax_c {p_{\w}}(\x)[c] \\
    0, & otherwise,
\end{cases}
\end{equation}
where $\Tilde{\y}_c$ is the pseudo label and ${p_{\w}}(\x)[c]$ is the model prediction value  of sample $\x$ for class $c$. 
\begin{algorithm}
\SetAlgoLined

\textbf{Input}:$\tD_l^k$ set of labeled samples in client $k$, set of unlabeled samples $\tD_u^k$, model agreement number $u$, dictionary $P$ that stored all model parameters \;\\
\For{each client $i \in S$ in parallel}{
    U := $\tD_u^k$ \;\\
    L := $\tD_l^k$ \;\\
    % $S =$ random sets of all K clients\;
    % $w_i^t = w_t$ \;
    \For{$\x \in U$}{
     \For {each model $p_{\w}^k$ }{
     $\Tilde{\y}^k \leftarrow $ using equation (\ref{conf_th}) to predict on $\x$
     }
     
     $s_x \leftarrow$ the number of most frequent element in local model predictions \;\\
     \If{$s_x \geq u$ }{
        U := $\tD_l^k \cup (\x,\Tilde{\y_i})$, where $\Tilde{\y_i}$ is defined by equation (\ref{equ:max_y_i})\\
        L := $\tD_l^k \setminus (\x, \y)$
        
    }

    }

} 
 \caption{\textbf{Pseudo-Labeling Procedure}}
 \label{Algorithm_pseudo_labeling}
\end{algorithm}

In our setting, each client in communication round $t$ has labeled data $\tD_{l_t}^k := \{(\x_i,\y_i)\}_{i=1}^{{N_l}_t}$, unlabeled data $\tD_{u_t}^k := \{\x_j\}_{j=1}^{{N_{u}}_t}$ and local model $p_{\w}^k$, $t \in {1,..,T}$. When the subset of unlabeled data $\Tilde{\tD}_{u_t}^k$
satisfies a predefined criterion, as shown in the following paragraph, the model will generate pseudo label as shown in equation \ref{eqs12}. Thus, the new labeled data and unlabeled data are generated as $\tD_{l_{t+1}}^k = \tD_{l_t}^k \cup \Tilde{\tD}_{u_t}^k$, $\tD_{u_{t+1}}^k =\tD_{u_t}^k \setminus \Tilde{\tD}_{u_t}^k$. In this way, the model propagates hard pseudo-labels (i.e. one-hot vectors) to the unlabeled data. We train the local model $p_{\w}^k$ employing standard cross entropy loss to both annotated true labels and pseudo-labels equally. 

The criteria used to select how many and which samples are transferred from unlabeled data to the labeled data during training at each round is key to our method. Model predictions that generated these hard labels are not all correct. It is important to reduce the error rate of pseudo-labels so that the error does not get propagated into future training. Previous studies have explored different types of uncertainty of the prediction to alleviate this problem, including using the highest confidence to label data \cite{zhu2005semi}, or performing label propagation based on the nearest feature space \cite{shi2018transductive}. We adopt the former approach but additionally use multiple clients to help with pseudo-label in our federated learning setting. At communication round $t$, the server broadcasts not only the aggregated model but also K other local models to each client. Thus, each client would have K+1 models stored locally and generate pseudo-labels based on multiple model agreements. Consider $\x_i$ as an unlabeled data point. $\x_i$ is fed to these K+1 models in parallel, where we can get $\Tilde{\y}_i^k = \mathds{1} ({p_{\w}^k}(\x) ) $ with $\mathds{1}(\cdot)$ producing one-hot labels with given softmax values. To maintain stability, in each client, we only retain artificial labels whose largest class probability is above a predefined threshold $\gamma_t$ for communication round $t$ \cite{lee2013pseudo}. 
\begin{equation}
\label{conf_th}
\Tilde{\y}_i^k = \mathds{1} ({p_{\w}^k}(x) \geq  \gamma_t).
\end{equation}

% \textcolor{red}{the following equ may have some problem}

In addition, we incorporate an unlabeled data point based on how many models agree on the same label. To generate a pseudo-label for data sample, we have 

\begin{equation}
\Tilde{\y_i}  = \text{Mode}(\y_i^1,\y_i^2...,\y_i^{K+1} ),  
\label{equ:max_y_i}
\end{equation}
where Mode$(\cdot)$ outputs the class category that has maximum model agreement. We then only keep the label if enough local models agree on the sample label (i.e. using a predefined model agreement number $u$, $u \leq \text{K}+1$). Algorithm \ref{Algorithm_pseudo_labeling} shows the pseudo-labeling procedure. The output of the algorithm is the updated local labeled and unlabeled datasets. The entire training procedure is presented in Algorithm \ref{alg.SemiFed}.

\section{Experimental Evaluation}
\label{experimenet_section}

\begin{figure*}[h]

  \includegraphics[width=.48\linewidth]{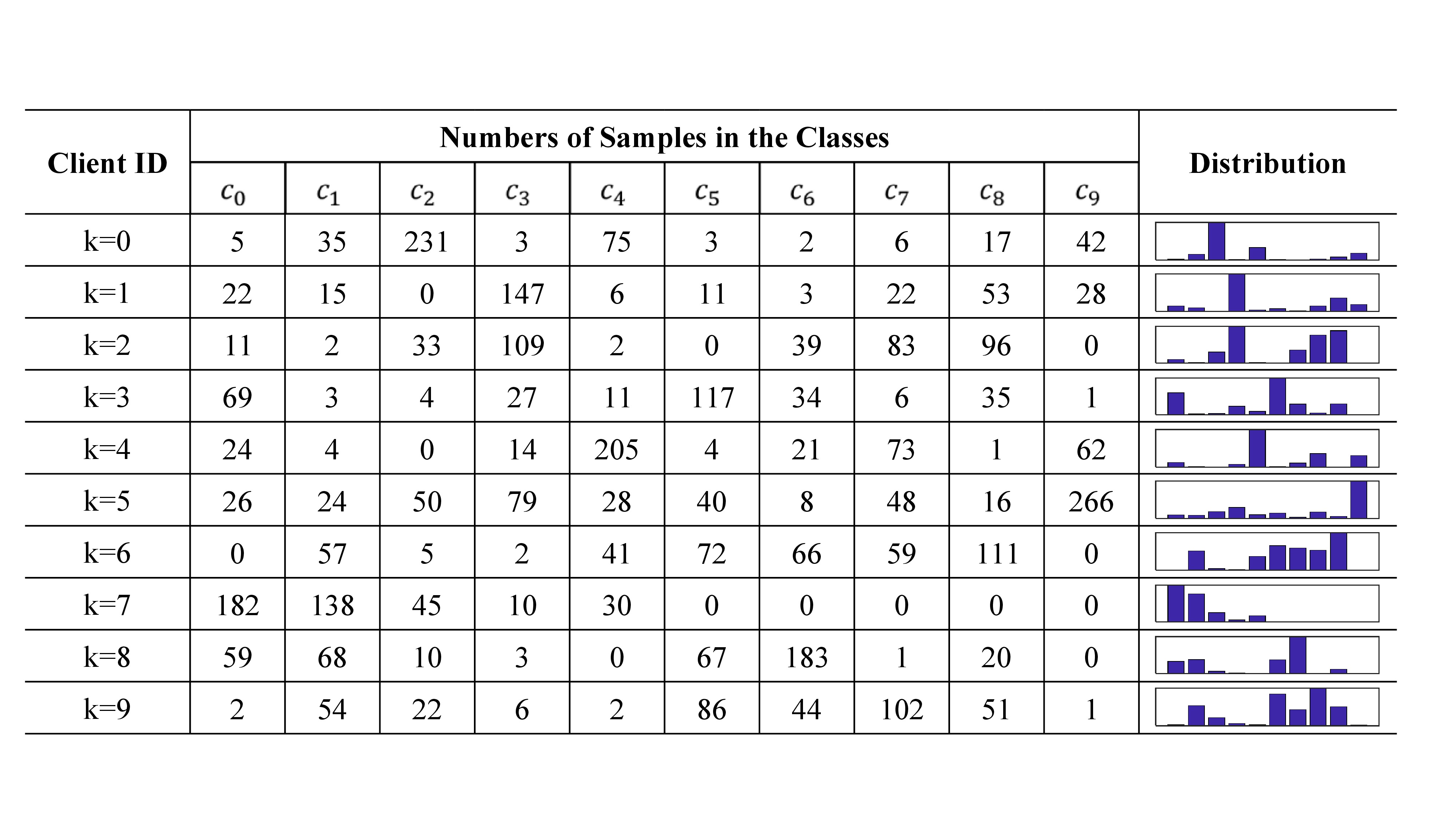}\hfill
\includegraphics[width=.48\linewidth]{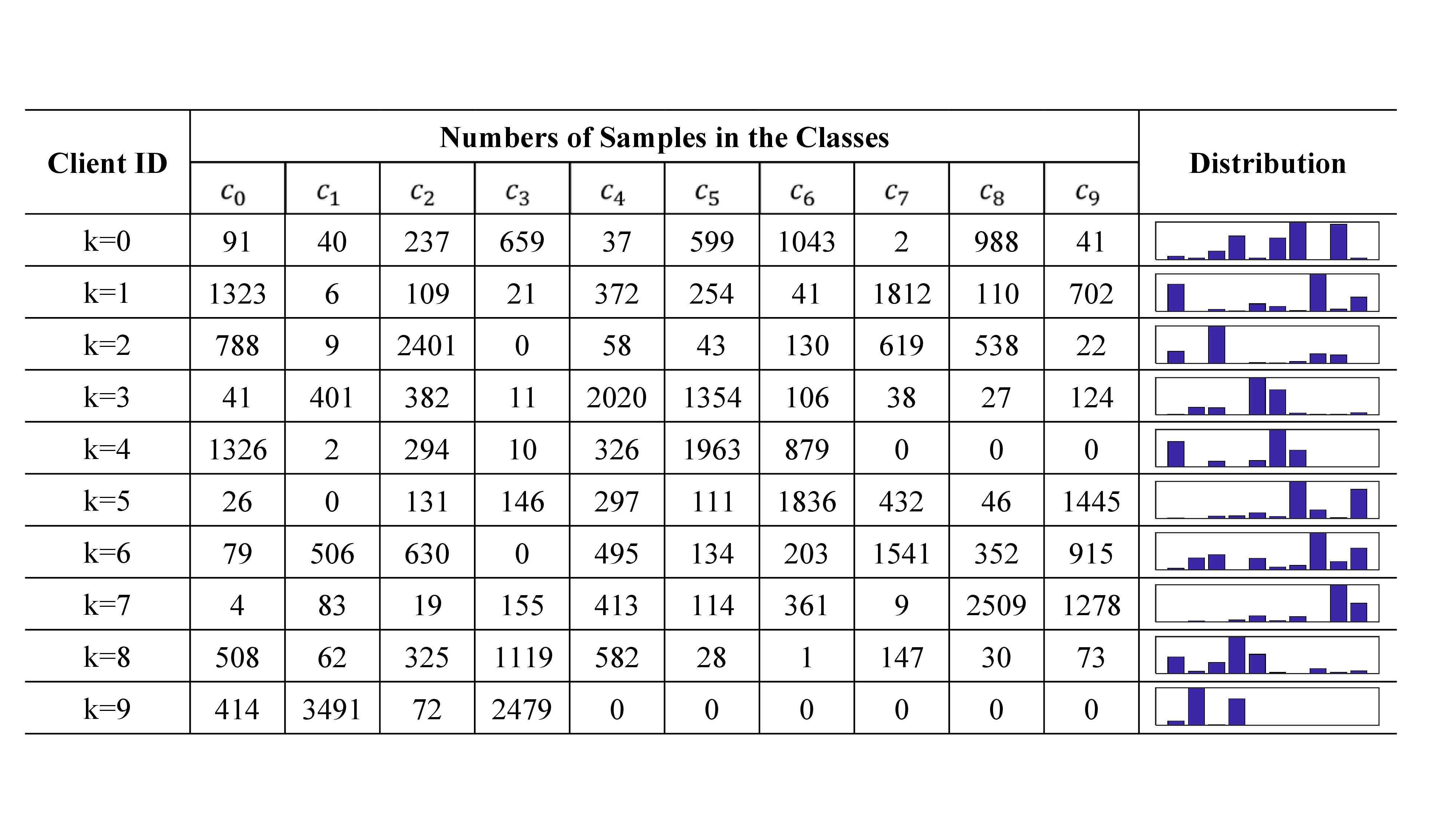}
  \caption{The distribution of non-IID data for CIFAR-10. The left table shows the data distribution for 4000 labeled images and the right table shows the data distribution for 46000 labeled images}
\label{hetero_data_partition}
\end{figure*}

\begin{table*}[ht]

\caption{Accuracy of federated learning with varying number of labeled sample on CIFAR-10 and SVHN using ResNet-50 architecture under IID data partition settings.
}
\centering
\begin{tabular}{c|c|c|c|c|c|c}
\hline
 & \multicolumn{3}{|c|}{CIFAR-10} & \multicolumn{3}{|c}{SVHN} \\
\hline
Method & 1000 labels & 2000 labels & 4000 labels & 1000 labels & 2000 labels & 4000 labels \\
% cell 1 & cell 2 & cell 3\\
% cell 1 & cell 2 & cell 3\\
\hline
Supervised & 46.97  & 60.31  & 66.57 & 49.17 & 82.93 & 86.43 \\
VAT        & 50.82 & 67.10   & 80.04 & 88.70  & 92.48 & 93.79 \\ 
UDA-consistency & 50.13  & 68.25  & 77.68 & 85.35 & 92.45 & 94.80 \\

SemiFed & 52.63  & 72.98  & 81.34 & 89.50 & 94.43 & 94.45 \\

\hline
Fully-Supervised &\multicolumn{3}{|c}{91.03} &\multicolumn{3}{|c}{94.81} \\
\hline
\end{tabular}
\label{result_homo}
\end{table*}

\begin{table*}[ht]

\caption{Accuracy of federated learning with varying the number of labeled samples on CIFAR-10 and SVHN using ResNet-50 architecture under non-IID data distribution settings.
}
\centering
\begin{tabular}{c|c|c|c|c|c|c}
\hline
 & \multicolumn{3}{|c|}{CIFAR-10} & \multicolumn{3}{|c}{SVHN} \\
\hline
Method & 1000 labels & 2000 labels & 4000 labels & 1000 labels & 2000 labels & 4000 labels \\
% cell 1 & cell 2 & cell 3\\
% cell 1 & cell 2 & cell 3\\
\hline
Supervised & 40.26  & 56.87  & 61.31 & 50.62 & 80.00 & 84.82 \\
VAT        & 45.19  &  60.26  & 67.85 & 83.76  & 90.27 & 93.14 \\ 
UDA-consistency & 44.29  & 58.24  & 71.57 & 87.88 & 91.42 & 93.30 \\

SemiFed & 46.46  & 62.63  & 75.41 & 89.50 & 92.81 & 94.01 \\
\hline
Fully-Supervised  &\multicolumn{3}{|c}{88.90} &\multicolumn{3}{|c}{92.99} \\
\hline
\end{tabular}
\label{result_hetero}
\end{table*}

\subsection{Experiment Setup}

In this section, we quantitatively evaluate
our algorithm with extensive experiments
on two image classification benchmarks.

 \emph{1) Dataset.} We evaluate our framework on two standard image classification datasets: CIFAR-10 \cite{cifar10} and Street View House Numbers (SVHN) \cite{netzer2011reading}. Both datasets have resolution 32×32. In each dataset, we keep a small portion of the training images labeled and leave the rest of the dataset unlabeled. We evaluate the performance on the independent test set. The details are as follows.

\noindent\textbf{CIFAR-10} contains 10 classes with 50K color images for training and 10K for testing. As recommended by \cite{oliver2018realistic}, we use a validation set of 5K samples for CIFAR-10 for choosing hyper-parameters and add it back to the training set to evaluate test accuracy. We report results using 1K, 2K, and 4K labeled samples.

\noindent\textbf{SVHN} has 73257 images for training and 26032 for testing. Similarly, we perform experiments with a varying number of labeled images $N_l =$ 1K, 2K, 4K. The test images are used for a global test after each round. 

 \emph{2) Non-IID data preparation.} In practical federated learning settings, we often encounter non-IID data at different clients. In the experiments, we study the case when different clients $i$ and $j$ have different data distribution $P_i$ and $P_j$, which follows the practice of many prior works \cite{hsieh2019non,mcmahan2017communication}. We report results on both IID and non-IID settings. We generate non-IID data partitions of each dataset by splitting training samples into 10
unbalanced partitions. Following the scheme in \cite{chaoyanghe2020fedml,yurochkin2019bayesian}, we generate the non-IID portion by sampling $p_c \sim  Dir_J (0.5)$, which is Dirichlet distribution with concentration parameter 0.5 and allocating a $p_{c,k}$ proportion of the training samples of class $c$ to local client $k$. The actual heterogeneous data distribution is presented in Figure \ref{hetero_data_partition}. Note that we do not use currently released federated learning benchmark datasets such as FEMNIST because they are too ``easy'' to classify, i.e., the model trained on a small subset of labeled data can already achieve high performance.

\begin{algorithm}
\caption{\textbf{SemiFed}, $|\tB_l|$, $|\tB_u|$ is the local minibatch size for labeled and unlabeled data, $T$ is the number of communication rounds, and $\eta$ is the learning rate, $E$ is the number
of local epochs, $I$ is the number of local iterations within each epoch, $\lambda_u$ is the coefficient for unsupervised loss, $T_p$ is the set of communication rounds that perform pseudo-labeling}
\SetAlgoLined
%\begin{algorithmic}[1]
\textbf{Server executes:} \; \\
Initialize $\w_0$; \;

 \For{each round $t=0 ,...,T-1$}{

    $\w^i_{t,0} = \w_t$;\; \\
    $P \leftarrow$  empty dictionary \; 
    
    \For{each client $k \in \tS$ in parallel}{
        %$\tB \leftarrow $ split local data into batches of size $B$ \; \\
        $\w^k_t \leftarrow$ ClientLocalTraining($k$) \;\\
        $P[k] = \w^k_t$ \;\\
    }
    $\w_{t+1} = \frac{1}{|\tS|} \sum\limits_{k \in \tS} \w^k_t$; \;\\
   \For{each client $k \in \tS$ in parallel}{
    \If{$t \in T_p$}{
        send model dictionary $P$ to client $k$ 
    }
    send aggregated model $\w_{t+1}$ to client $k$ 
  }

}

\textbf{ClientLocalTraining($k$):} \; \\
\For{local epoch $e$ from $0$ to $E-1$}{
    \For{iteration $i$ from $0$ to $I-1$}{
    Sample mini-batch $\tB_l$ from $\tens{D}_l^k$; \;\\
    Sample mini-batch $\tB_u$ from $\tens{D}_u^k$; \;\\
     // $ l_s$is computed using Eq. \ref{supervised_loss_equation} \;\\
     // $l_u$ is computed using Eq. \ref{unsupervised_loss_equation} \;\\
    $\w_{t,i+1}^k = \w_{t,i}^k - \eta \cdot\frac{1}{|\tB_l|}\sum _{b_l\in\tB_l}\nabla l_s(\w^k_{t,i};\x_{l,b},\y_b)- \lambda_u \cdot \eta \cdot\frac{1}{|\tB_u|}\sum _{b_u\in\tB_u}\nabla l_u(\w^k_{t,i};\x_{u,b})$
        }
    \If{$t \in T_p$}{
        Execute Algorithm \ref{Algorithm_pseudo_labeling}
    }

}
%\end{multicols}

%\end{algorithmic}
\label{alg.SemiFed}
\end{algorithm}

 \emph{3) Implementation details.} Unless otherwise specified, we use the following experiment implementation as the standard setting: we setup 10 clients and run on a GPU server for all datasets and models. We test our framework using
ResNet-50 \cite{he2016deep} as the base model architecture for the CIFAR-10 and SVHN datasets. There are several important hyper-parameters in our SemiFed framework: the number of total communication rounds, the edge-side epoch number, the unsupervised balancing parameter $\lambda_u$, the predefined confidence threshold $\gamma_t$ and the model agreement number $u$. After a grid search for tuning parameters, we run all experiments with 300 communication rounds, unsupervised ratio $\lambda_u = 1.0$, model agreement number $u=11$ (all local models and the global model have to agree on the same label for a single data point if we want to pseudo-label it). We will also show a detailed parameter study in Section 4.3. We apply pseudo-labeling method at communication round $t=50,100$ for SVHN and $t=50,100,200$ for CIFAR-10. For the confidence threshold, we set 0.98 for SVHN and IID data partitions of CIFAR-10 and $0.9,0.85,0.7$ for non-IID partitions of CIFAR-10. We choose the confidence threshold based on how many samples are given labels (up to 1000 per round). The client-side epoch number depends on the dataset. For CIFAR-10, we set it to be 10, and for SVHN we set it to be 5 since SVHN is much easier to classify. On the client-side, we use Stochastic Gradient Descent with nesterov momentum as the optimizer, where the momentum factor is 0.9, $L_2$ regularization is 0.0001, the constant learning rate is $lr= 0.3$, and the batch size is 64.

Current popular distributed training libraries in PyTorch \cite{paszke2019pytorch} and TensorFlow \cite{abadi2016tensorflow} (i.e. TensorFlow-Federated (TFF), PySyft \cite{theodblp}, and LEAF \cite{caldas2018leaf}) lack the support of our algorithm (pseudo-labeling with multiple model transferring). To simplify the usage of our algorithm, we build our algorithm on FedML \cite{chaoyanghe2020fedml}, an open-source library for federated learning and deploy it in a distributed computing environment. We conduct the experiments on a computer with three NVIDIA GeForce GTX 2080 GPUs to report the performance.

 \emph{4) Compared Baselines.} We evaluate the following approaches:

\begin{itemize}
  \item \textbf{Supervised.} Supervised refers to using only labeled
data points from CIFAR-10 and SVHN respectively, without
any unlabeled data on the client-side. The server still performs FedAvg and aggregates local models and broadcasts the aggregated global model to all clients.

\item \textbf{Fully Supervised}. we also train a fully supervised baseline (use all the original labels of the training samples) to measure the the ideal accuracy we could hope to obtain under federated learning settings.

\item \textbf{VAT.} VAT stands for Virtual Adversarial Training \cite{miyato2018virtual}. It applies a small perturbation to both the labeled and unlabeled input in the direction that can mostly alter the prediction of the model from the current inferred label by the model.

\item \textbf{UDA-consistency \cite{xie2019unsupervised}.} On the client-side, UDA uses consistency loss only to enforce the model being insensitive to the noise added to the input and hence smoother with respect to changes in the input space. It relies on advanced data augmentation\cite{cubuk2020randaugment} which is the same augmentation technique in our consistency regularization to add the noise. We re-implement this method in federated learning to give a fair comparison.

\end{itemize}

\subsection{Performance Comparison with Baselines}

We evaluate the algorithm on CIFAR-10 and SVHN
datasets using ResNet-50 and report our results under IID and non-IID setting in Table \ref{result_homo} and Table \ref{result_hetero}, respectively. The relative performance against the baseline approaches vary across different datasets with a different number of labeled samples. Clearly, our algorithm outperforms the supervised method that trains the network using only labeled samples by a large margin. We find that even using consistency loss alone can improve model performance a lot compared with the model trained with only labeled data, demonstrating the advantage of consistency-regularization based method exploiting unlabeled information. Note that under SVHN non-IID data distribution with our consistency regularization and pseudo-labeling method, we achieve accuracy better than using all labeled data without advanced data augmentation. Overall, our approach improves the test accuracy with an improvement raging from 0.71\% to 4.39\% improvement on CIFAR-10 and matches the test accuracy on SVHN when compared with all previous methods that only rely on consistency loss \cite{xie2019unsupervised}.

\subsection{Parameter Study}
% Since hyper-parameter tuning is significant in deep learning, we perform an extensive  to better 
Deep neural networks trained for image classification can be heavily affected by the architecture, training schedule, etc. Moreover, since our algorithm is built upon the assumption where decision boundary should lie in low-density regions to improve generalization performance, other factors such as the $L_p$ regularization term should be significant. Thus, we want to quantitatively show the importance of these factors and understand the effects of different parameter settings of our algorithm on the final results. We focus on studying with a single 4000 label split from CIFAR-10 under non-IID data partitions and report results using constant learning rate $lr =0.3$ for consistency regularization.

 \emph{1) Effect of the unsupervised ratio $\lambda_u$.}
 Parameter $\lambda_u$ controls the weight balance between the labeled data and the unlabeled data loss. $\lambda_u$ being too small would cause the model to be mainly controlled by supervised loss and to learn nothing from unlabeled data. On the other hand, if $\lambda_u$ is too large, the model would neglect useful guidance from labels. We illustrate the effect
of varying unsupervised ratio on the validation performance in Figure 3a. We found that both $\lambda_u =1.0$ and $\lambda_u =2.0$ yields good performance. Following the setting in \cite{miyato2018virtual}, we choose $\lambda_u =1.0$ in all our experiment.

\begin{figure}
\centering
\begin{minipage}{.245\textwidth}
% \begin{subfigure}[b]{0.245\textwidth}
\includegraphics[width=\textwidth]{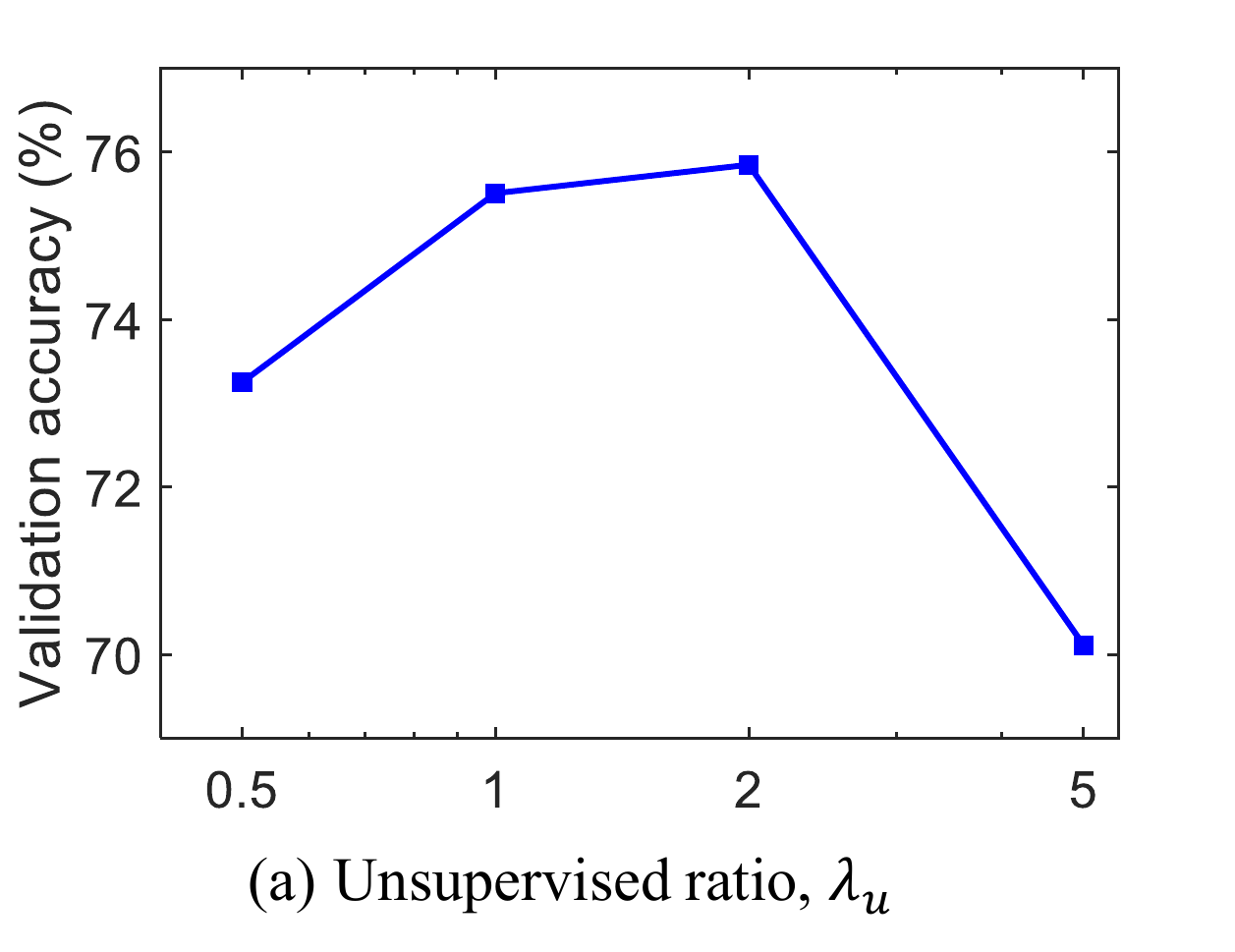} 

% \end{subfigure}
\label{fig:unsupervised_ratio}
\end{minipage}
\hspace{-1em}
\begin{minipage}{.245\textwidth}
% \begin{subfigure}[b]{0.245\textwidth}
\includegraphics[width=\textwidth]{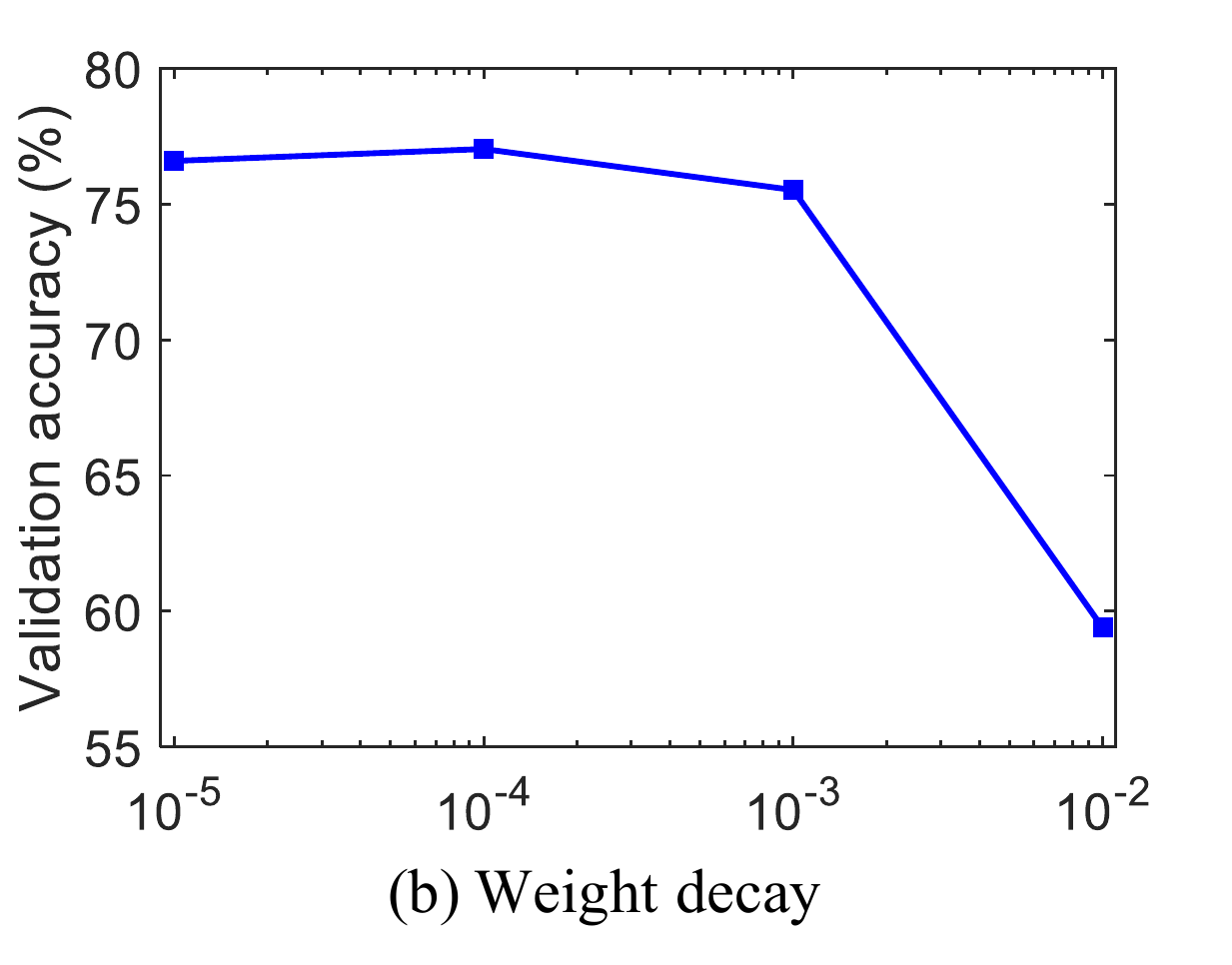}
\label{fig:subim2}
% \end{subfigure}

\label{fig:image2}
\end{minipage}
\caption{ Plots of various parameter studies on SemiFed (a) Varying the ratio of unsupervised loss (b) Varying the loss coefficient for weight decay.}
\end{figure}

\emph{2) Effect of the $L_2$ regularization in client-side.} 
  We apply $L_2$ regularization in all our experiment. $L_2$ regularization is particularly important since it can have an effect of mitigating the over-sensitivity of the output with respect to input. Figure 3b. demonstrates the effect of choosing the loss coefficient for $L_2$ regularization. We find that choosing a large coefficient can degrade performance by five percent.

\emph{3) Effect of the learning rate in client-side.} In Table \ref{learning rate}, we compare the training curves with different learning rates and learning rate decay schedules. We find that the learning rate can have a substantial effect on network performance. Given an inappropriate learning rate, i.e., $lr = 0.03$, the model incurs a proportionally large classification error. Moreover, a popular technique in recent works for image classification is to use adaptive learning rate decay schedules that can encourage fast convergence \cite{sohn2020fixmatch}. Thus, we try two different learning rate schedules: stagewise step decay schedule \cite{he2016deep} and cosine learning rate decay \cite{loshchilov2016sgdr}. However, we do not obtain performance gain by using learning rate decay schedules under our SemiFed framework. Thus, we maintain the constant learning rate in all our experiments.
\begin{table}[h!]
    \centering
       \caption{Parameter study on learning rate and decay schedules. Accuracy is reported on CIFAR-10 with 4000 labeled samples and non-IID data}
    \label{learning rate}
    \begin{tabular}{c | c}
    \hline
    Decay Schedule & Validation Acc (\%)  \\
    \hline
    constant $lr=0.3$ & 75.51  \\ 
    constant $lr=0.03$ & 69.54   \\
    stagewise decay & 73.14  \\
    cosine decay & 73.63\\
    \hline
    \end{tabular}
\end{table}

\subsection{Ablation Study}

In order to understand the influence of consistency loss and pseudo-labeling method individually in model performance, we perform an ablation study by isolating each procedure's effects.

\begin{figure}[h]

  \includegraphics[width=0.98\linewidth]{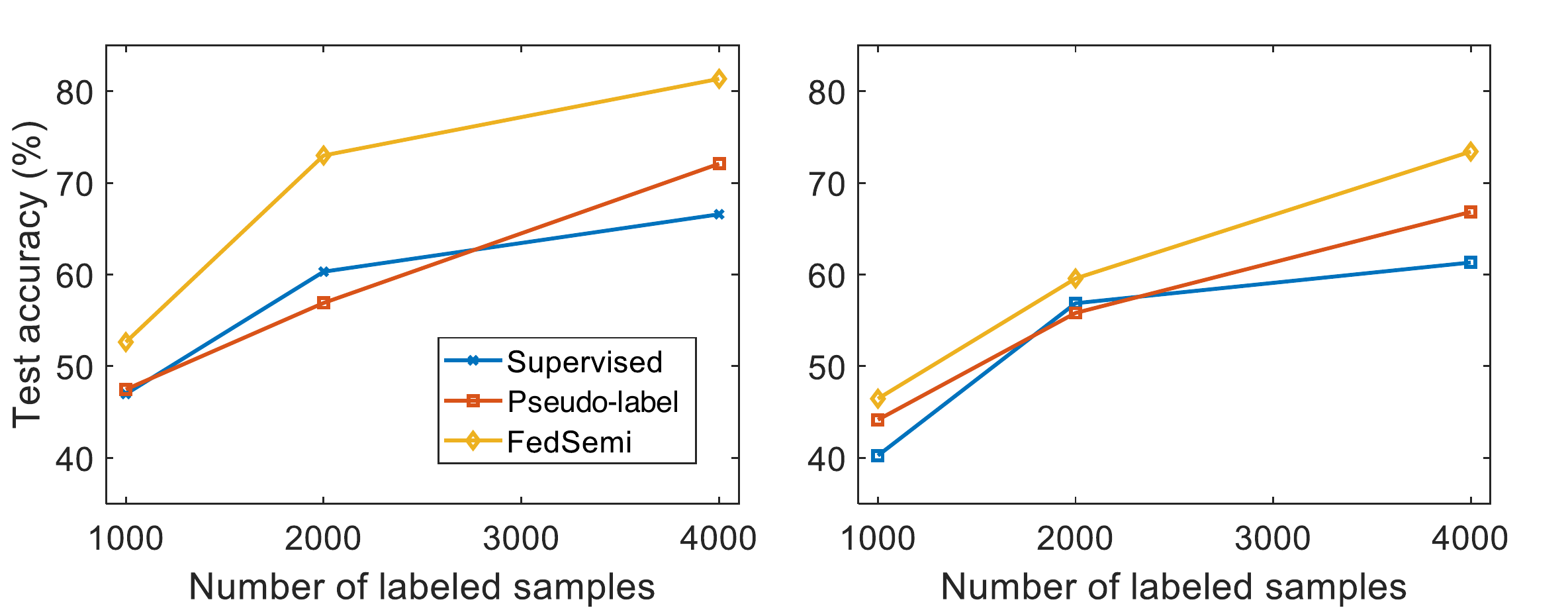}

  \caption{Test accuracy of using Supervised, Pseudo-labeling method without consistency loss and SemiFed algorithm on CIFAR-10. The left image shows the test accuracy under IID data partitions and the right image shows the test accuracy under non-IID data partitions}
\label{pseudo_label_only}
\end{figure}

 \emph{1) Benefit of applying consistency regularization.} In this section, we are
interested in evaluating how consistency regularization can
benefit exploiting information from unlabeled data. To isolate the contribution of consistency regularization in the neural network, we train the model using only the labeled
data. Then, we apply the pseudo-labeling method to generate labels for unlabeled data to guide the learning process. Specifically, we compare it with the approach that only uses supervised data and our SemiFed algorithm. Figure \ref{pseudo_label_only} presents the results on CIFAR-10 under IID and non-IID settings.
 
\begin{figure}[h]

  \includegraphics[width=0.98\linewidth]{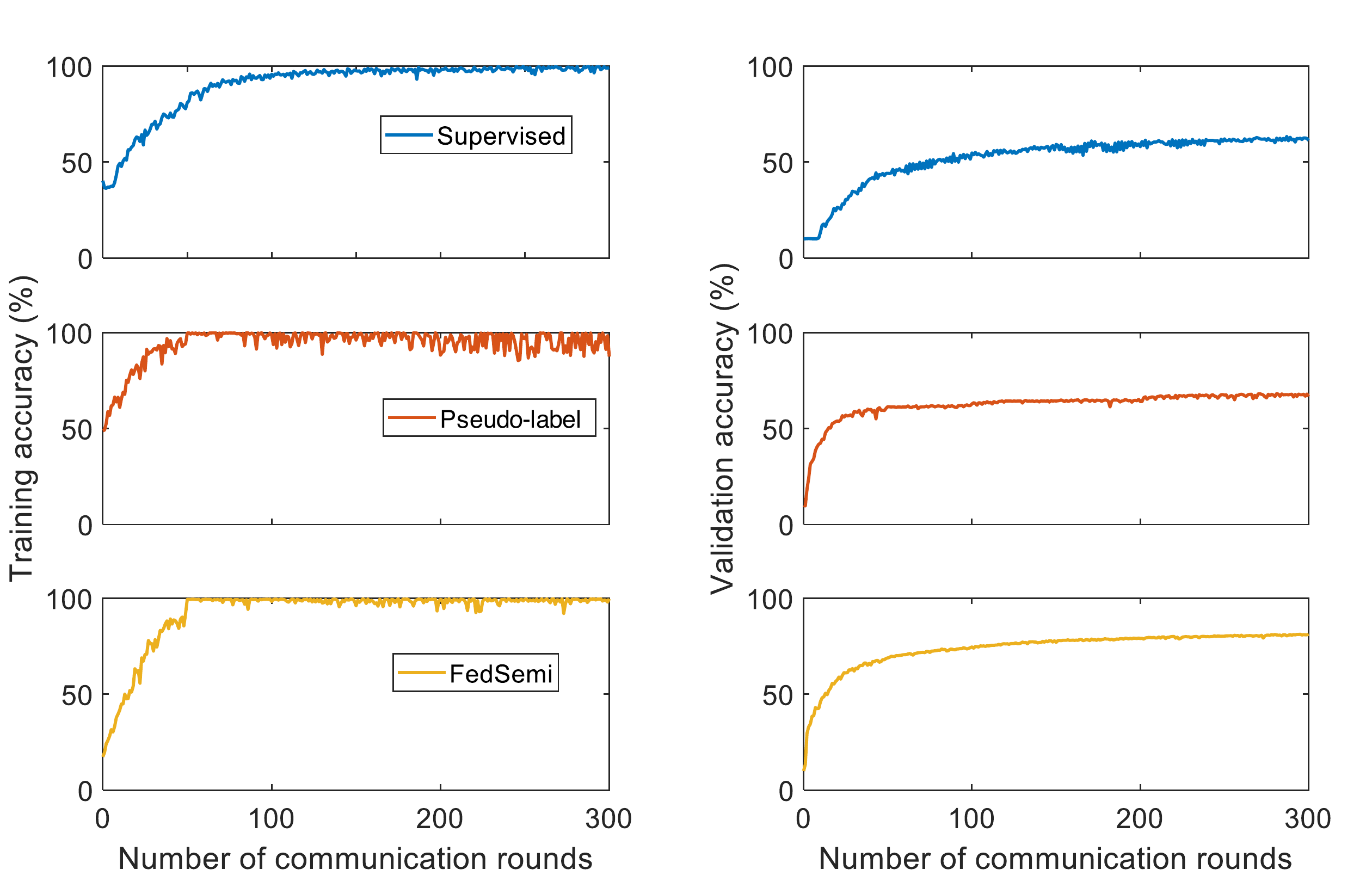}

  \caption{Training accuracy and validation accuracy of using Supervised, Pseudo-labeling method without consistency loss and SemiFed algorithm on CIFAR-10 with 4000 labeled images under non-IID data partitions}
\label{pseudo_label_valid_train}
\end{figure}

It is clear that consistency regularization plays a significant role in our model performance. To further understand the role of consistency regularization in network training, we present the validation accuracy and training accuracy (the accuracy is aggregated among all the clients) for CIFAR-10 with 4000 labeled samples in non-IID partitions (See Figure \ref{pseudo_label_valid_train}). From the curve we can make the following observations: compared with the method that applies pseudo-labeling without consistency loss, our algorithm has a smoother curve in training accuracy. This is because the pseudo-labeling method introduces noisy labels to the dataset which may hurt model performance. Meanwhile, the validation accuracy of the supervised method that only trains with 4000 labeled samples also shows that the trained model does not have good generalization accuracy. The consistency regularization helps the network to fit the training data perfectly and provide better generalization performance \cite{xie2019unsupervised,tarvainen2017mean}. 

 \emph{2) Benefit of pseudo-labeling.} We are interested in evaluating how the pseudo-labeling benefits learning performance in SemiFed. Since UDA uses the same augmentation as SemiFed without pseudo-labeling, we compare federated UDA baselines against our SemiFed algorithm. We note that our  method performs the best on all almost all experiments, verifying the benefit of pseudo-labeling. The only exception is for SVHN, where UDA is a bit superior with 4000 labeled data samples. This may be explained by the fact that image classification on SVHN is not a difficult task, and thus, we can get the trained model with relatively high accuracy on both the test set and the training set. As shown in the Table \ref{result_hetero}, the gap between the model trained on the fully supervised training samples and the SVHN with 4000 labeled data is small, and thus pseudo-labeling method cannot bring any performance gain to the network. In addition, we can also observe from Figure \ref{pseudo_label_only} that pseudo-labeling alone performs better than supervised approach in most cases, demonstrating the effectiveness of pseudo-labeling in the training procedure.

% \section{Related Work}
% \label{related_work_section}

\section{Conclusion}
\label{conclusion_section}
 We proposed SemiFed, a unified framework to address the challenge in limited samples for image classification based on a combination of consistency regularization and pseudo-labeling. We adapt pseudo-labeling method to FL setting by allowing multiple models to agree on the same dataset to improve robustness accuracy of the network. The experimental results confirm the feasibility of our conceptually sim- ple approach compared with baselines in two datasets under both IID and non-IID settings.

In the future, we plan to explore how to enhance synergies of pseudo-labeling and consistency regularization and design an automatic scheme for deciding the threshold or criteria used for adding training samples across communication rounds.

\bibliography{ref}
\bibliographystyle{siamplain}

\end{document}